\documentclass[10pt,twocolumn,letterpaper]{article}

\usepackage{wacv}
\usepackage{times}
\usepackage{epsfig}
\usepackage{graphicx}
\usepackage{amsmath}
\usepackage{amssymb}

\usepackage{multirow}
\usepackage{booktabs}
\usepackage{soul}
\usepackage{amsmath}
\DeclareMathOperator*{\argmax}{arg\,max}
\DeclareMathOperator*{\argmin}{arg\,min}

\usepackage{float}
\usepackage{subcaption}

\usepackage[table]{xcolor}
\newcommand{\myparagraph}[1]{\vspace{0pt}\noindent{\bf #1}}

\wacvfinalcopy % *** Uncomment this line for the final submission

\def\wacvPaperID{278} % *** Enter the wacv Paper ID here

\ifwacvfinal\fi
\begin{document}

%%%%%%%%% TITLE
\title{Unsupervised Adaptation for Synthetic-to-Real Handwritten Word Recognition}

% Authors at the same institution
%\author{Lei Kang, Mar\c{c}al Rusi{\~n}ol, Alicia Forn\'{e}s, Pau Riba \\
%Computer Vision Center, Universitat Aut{\`o}noma de Barcelona\\ Barcelona, Spain\\
%{\tt\small \{lkang, marcal, afornes, priba\}@cvc.uab.es}
%\and
% Authors at different institutions
%Mauricio Villegas \\
%omni:us,\\ Berlin, Germany\\
%{\tt\small mauricio@omnius.com}
% \and
% Second Author \\

\author{Lei Kang$^{* \dag}$, Mar\c{c}al Rusi{\~n}ol$^{*}$, Alicia Forn\'{e}s$^{*}$, Pau Riba$^{*}$, Mauricio Villegas$^{\dag}$ \\
$^{*}$Computer Vision Center, Universitat Aut{\`o}noma de Barcelona, Barcelona, Spain\\
{\tt\small \{lkang, marcal, afornes, priba\}@cvc.uab.es}
\\
$^{\dag}$omni:us, Berlin, Germany\\
{\tt\small \{lei, mauricio\}@omnius.com}
}

%\author{
%    \IEEEauthorblockN{Lei Kang\IEEEauthorrefmark{1}\IEEEauthorrefmark{2}, Mar\c{c}al Rusi{\~n}ol\IEEEauthorrefmark{1}, Alicia Forn\'{e}s\IEEEauthorrefmark{1}, Pau Riba\IEEEauthorrefmark{1}, Mauricio Villegas\IEEEauthorrefmark{2}}\\
%    \IEEEauthorblockA{\IEEEauthorrefmark{1}Computer Vision Center, Universitat Aut{\`o}noma de Barcelona, Barcelona, Spain\\\{lkang, marcal, afornes, priba\}@cvc.uab.es}\\
%    \IEEEauthorblockA{\IEEEauthorrefmark{2}omni:us, Berlin, Germany\\\{lei, mauricio\}@omnius.com}
%}

\maketitle
\ifwacvfinal\thispagestyle{empty}\fi

%%%%%%%%% ABSTRACT
\begin{abstract}
Handwritten Text Recognition (HTR) is still a challenging problem because it must deal with two important difficulties: the variability among writing styles, and the scarcity of labelled data. To alleviate such problems, synthetic data generation and data augmentation are typically used to train HTR systems. However, training with such data produces encouraging but still inaccurate transcriptions in real words. In this paper, we propose an unsupervised writer adaptation approach that is able to automatically adjust a generic handwritten word recognizer, fully trained with synthetic fonts, towards a new incoming writer. We have experimentally validated our proposal using five different datasets, covering several challenges (i) the document source: modern and historic samples, which may involve paper degradation problems; (ii) different handwriting styles: single and multiple writer collections; and (iii) language, which involves different character combinations. Across these challenging collections, we show that our system is able to maintain its performance, thus, it provides a practical and generic approach to deal with new document collections without requiring any expensive and tedious manual annotation step.
\end{abstract}

%%%%%%%%% BODY TEXT
\section{Introduction}

\begin{figure}
\centering
\begin{tabular}{ccc}
\toprule
\includegraphics[width=2.3cm,height=.75cm]{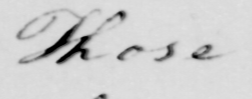}&
\includegraphics[width=2.3cm,height=.75cm]{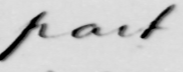}&
\includegraphics[width=2.3cm,height=.75cm]{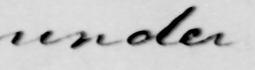}\\
\small{\texttt{Hoss}} & \small{\texttt{fasl}} &\small{\texttt{werder}}\\
$\downarrow$& $\downarrow$& $\downarrow$\\
\small{\texttt{\textbf{those}}} & \small{\texttt{\textbf{part}}}  & \small{\texttt{\textbf{under}}}\\
\midrule
\includegraphics[width=2.3cm,height=.75cm]{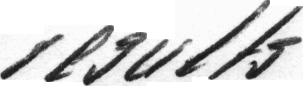}&
\includegraphics[width=2.3cm,height=.75cm]{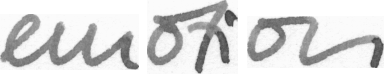}&
\includegraphics[width=2.3cm,height=.75cm]{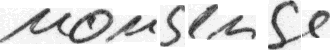}\\
\small{\texttt{REIUIL}} & \small{\texttt{eUOROV}}  & \small{\texttt{MONSEASe}}\\
$\downarrow$& $\downarrow$& $\downarrow$\\
\small{\texttt{\textbf{results}}} & \small{\texttt{\textbf{emotion}}} & \small{\texttt{\textbf{nonsense}}}\\
\midrule
\includegraphics[width=2.3cm,height=.75cm]{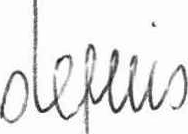}&
\includegraphics[width=2.3cm,height=.75cm]{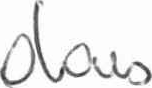}&
\includegraphics[width=2.3cm,height=.75cm]{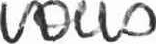}\\
\small{\texttt{oleuus}} & \small{\texttt{olous}}  & \small{\texttt{1000}}\\
$\downarrow$& $\downarrow$& $\downarrow$\\
\small{\texttt{\textbf{depuis}}} & \small{\texttt{\textbf{dans}}} & \small{\texttt{\textbf{vous}}}\\
\midrule
\includegraphics[width=2.3cm,height=.75cm]{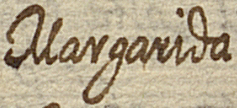}&
\includegraphics[width=2.3cm,height=.75cm]{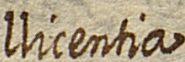}&
\includegraphics[width=2.3cm,height=.75cm]{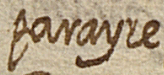}\\
\small{\texttt{Ilargarivia}} & \small{\texttt{lhicenhar}}  & \small{\texttt{favanye}}\\
$\downarrow$& $\downarrow$& $\downarrow$\\
\small{\texttt{\textbf{Margarida}}} & \small{\texttt{\textbf{llicentia}}} & \small{\texttt{\textbf{parayre}}}\\
\midrule
\includegraphics[width=2.3cm,height=.75cm]{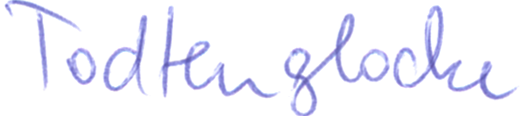}&
\includegraphics[width=2.3cm,height=.75cm]{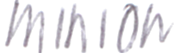}&
\includegraphics[width=2.3cm,height=.75cm]{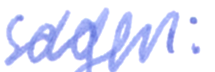}\\
\small{\texttt{Todfenglodu}} & \small{\texttt{MihIOw}}  & \small{\texttt{SIGIN}}\\
$\downarrow$& $\downarrow$& $\downarrow$\\
\small{\texttt{\textbf{Todtenglocke}}} & \small{\texttt{\textbf{minion}}} & \small{\texttt{\textbf{sagen}}}\\
\bottomrule
\end{tabular}
\caption{Handwritten word recognition results with our model trained only using synthetically generated word samples. We show the transcription before and after (in boldface) the unsupervised writer adaptation, for the GW, IAM, RIMES, Esposalles and CVL datasets respectively.}\label{fig:qualitative_samples}
\end{figure}

Handwritten Text Recognition (HTR) is a difficult task to automate by means of computer vision and machine learning techniques, mainly because of both the inter- and intra-class variability. Different instances of the same word, written by different people, will inevitably be composed by a succession of rather different glyphs, and thus, will end up looking very disparate from one sample to another. In the same sense, the same character written by the same writer, might look very different depending on the context when it was written. We humans, once we learn how to read scripted words, perform quite well at reading handwritten texts produced by individuals with handwriting styles that we have never seen before. However, computational models strive at being so generic unless they are supplied with huge amounts of training data coming from many different writers. 

But, gathering such huge annotated collections of training data quickly becomes too expensive. Although in the literature some publicly available benchmarking datasets have been established, such as IAM~\cite{marti2002iam} or RIMES~\cite{augustin2006rimes}, their volumes are still far away from nowadays large scale datasets like ImageNet or Open Images V5, that contain millions of annotations. Without such large amount of training data, deep learning architectures for HTR are prone to overfit to the seen writers and not generalize well. In order to cope with such lack of training data, on the one hand, some authors propose to engineer realistically looking data augmentation techniques~\cite{varga2008perturbation,Wigington17,poznanski2016cnn}, so that the amount of samples in the training dataset grows exponentially. On the other hand, an even cheaper and ever-growing strategy is to use fully synthetic training sets. Truetype electronic fonts that are designed with calligraphic styles are used to render randomly selected word images. Such approaches have been successfully leveraged to pre-train both scene and handwritten text recognition and retrieval systems~\cite{jaderberg2014synthetic, ahmad2015training, krishnan2018hwnet}. However, even if the synthetic fonts are carefully selected, the extracted visual features will most likely differ from the ones one might find when dealing with real handwritten text. In that sense, a final adjustment step is needed in order to bridge the representation gap between the synthetic and the real samples.

Such issue raised awareness of the document analysis community, that has researched on the topic of writer adaptation since the early nineties~\cite{matic1993writer,platt1997constructive,gilloux1994writer}. The main motivation of such applications, consist in adapting a generic writer-independent HTR system, trained over a large enough source dataset, towards a new distribution of a particular writer. Especially interesting are the approaches that ara able to yield such writer adaptation step in an unsupervised manner, that is, without needing any ground-truth labels from the new target writer.

Our main application contribution stems for the use of unsupervised domain adaptation to forge an annotation-free handwriting recognition system. Our proposed approach is fully trained with synthetically generated samples that mimic the specific characteristics of handwritten text. Later, it is unsupervisedly adapted towards any new incoming target writer. In particular, the system produces transcriptions (without the need of labelled real data) that are competitive even compared to supervised methods. Text being a sequential signal, several temporal pooling alternatives are proposed to redesign current domain adaptation techniques so that they are able to process variable-length sequences. All in all it represents a step towards the practical success of HTR in unconstrained scenarios.

We show some examples of the results obtained after such writer adaptation in Fig.~\ref{fig:qualitative_samples}. We observe that even though the synthetically trained model outputs gibberish text, the committed errors are quite understandable, since the confusion is between letters and glyphs that are visually close. Once the unsupervised writer adaptation is applied, the text is correctly transcribed in all those cases. Our proposal is validated by using five different datasets in different languages, showing that our handwritten word recognizer is adapted to modern and historic samples, single and multi-writer collections. Our proposed adaptable handwritten word recognition model outperforms the state of the art, and compares quite favourably to supervised fine-tuning methods while not needing any manually annotated label.

%%%%%%%%%%%%%%%%%%%%%%%%%%%%%%%%%%%%%%%%%%%%%%%%%%%%%%%%%%%%%%%%%%%%%%
%%%%%%%%%%%%%%%%%%%%%%%%%%%%%%%%%%%%%%%%%%%%%%%%%%%%%%%%%%%%%%%%%%%%%%
\section{Related Work}

\begin{figure*}[h!tb]
\includegraphics[width=2\columnwidth]{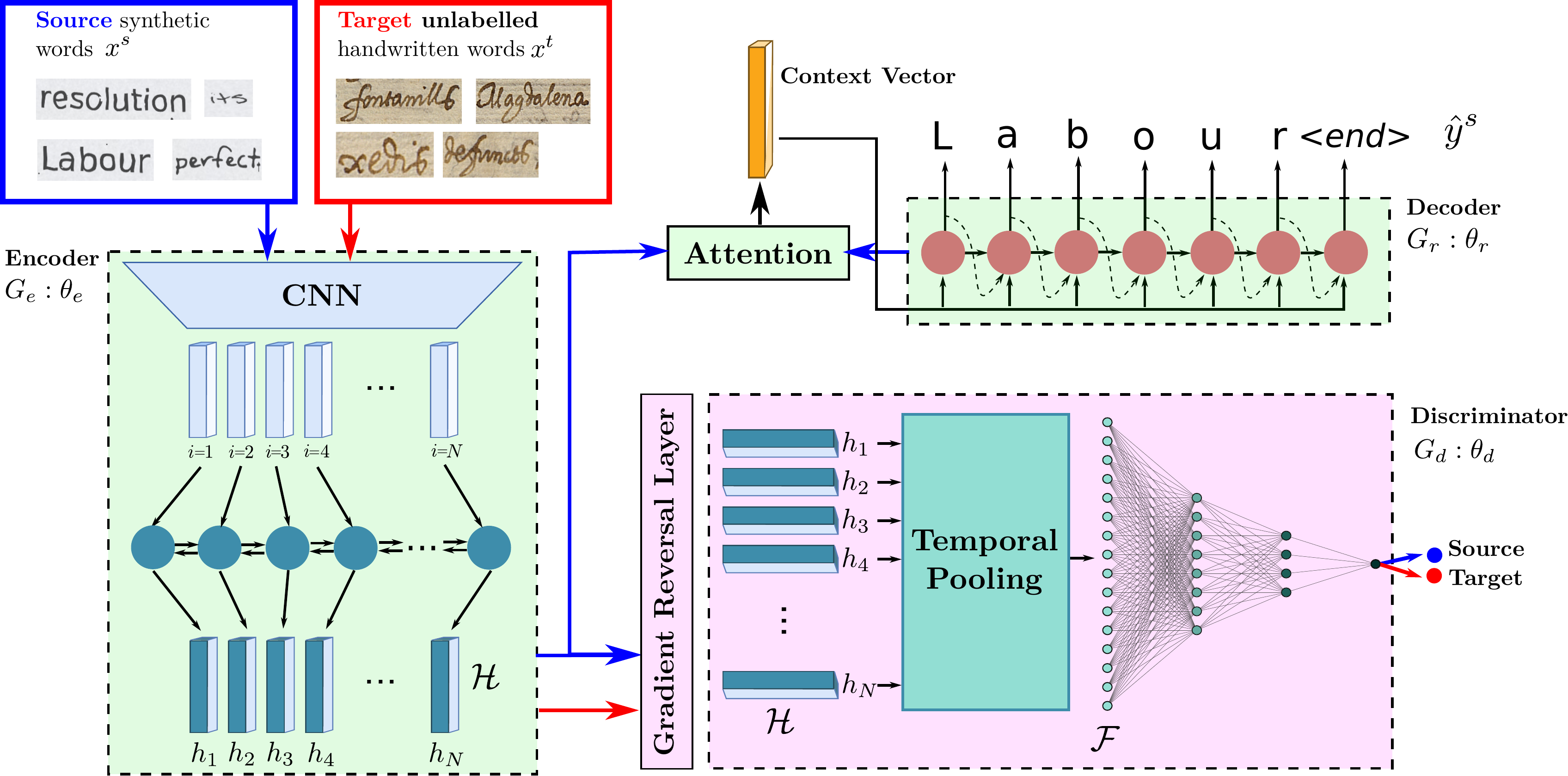}
\caption{Architecture of the adaptable handwritten word recognizer. The model consists of an encoder, a decoder and a discriminator. The discriminator incorporates a temporal pooling step to be able to adapt to variable-length sequences. The blocks corresponding to the handwriting recognizer, and therefore used for inference, are highlighted in light green; the block responsible for the unsupervised domain adaptation during training is highlighted in light magenta (best viewed in color).}
\label{fig:arch}
\end{figure*}

Inspired by the speech recognition community, writer adaptation techniques have been applied to modify early handwritten text recognition models based on \emph{Hidden Markov Models} ~\cite{gilloux1994writer,rodriguez2010unsupervised,ahmad2015training}. Once an omni-writer model has been trained, the model parameters, consisting of the Gaussian mixture means and variances, can be modified to better fit the target data distribution. Other early works proposed an \emph{Expectation-Maximization} strategy~\cite{nosary2004unsupervised,szummer2006discriminative} over a set of different character recognizers. The main advantage of such techniques was that the adaptation procedure to unseen target writers was done in an unsupervised manner, without needing any target labelled data.

With the rise of deep learning, the use of \emph{Long Short-Term Memory} (LSTMs) architectures became established for HTR. Such data hungry approaches have been commonly trained with the largest publicly available datasets, and then fine-tuned to the target collection to be recognized. Such tuning strategies~\cite{aradillas2018boosting,granet2018transfer,nair2018knowledge} guarantee that the neural networks can be properly trained, ending up extracting relevant features from handwriting strokes, that are later revamped to the target collection. But fine-tuning presents the downside of needing manual annotations both from the source and target datasets. In order to alleviate such pain, the use of synthetically generated texts as source data has lately surfaced ~\cite{krishnan2018hwnet,gurjar2018learning,bhunia2019handwriting}. By the use of synthetic fonts, overfitting is avoided at no labelling cost. However, HTR models fully trained on synthetically generated data still need to be grounded with real data in order to be effective, and thus target labels are still needed.

In order to discard target labelled data, unsupervised domain adaptation techniques have been proposed in the literature. Given a labeled source dataset and an unlabeled target dataset, their main goal is to adjust the recognition model so that it can generalize to the target domain while taking the domain shift across the datasets into account. A common approach to tackle unsupervised domain adaptation is through an adversarial learning strategy~\cite{ganin2014unsupervised,ganin2016domain,peng2018zero,tzeng2017adversarial}, in which the discrepancy across different domains is minimized by means of jointly training a recognizer network and a domain discriminator network. The recognizer seeks to correctly recognize the labeled source domain data, whereas the discriminator has to distinguish between samples drawn either from source or target domains. The adversarial model is trained jointly in a min-max fashion, in which the aim is to minimize the recognition loss while maximizing the discriminator loss. For instance, Ganin \etal~\cite{ganin2016domain} adapted a digit recognizer trained on handwritten digits from MNIST to tackle other target digit datasets such as MNIST-M or SVHN; or Yang \etal~\cite{yang2018deep}, who proposed an unsupervised domain adaptation scheme for Chinese characters across different datasets. Such strategy has been proven to be effective when dealing with classification problems, where the source and target domains share the same classes. However it can not be straightforwardly applied to HTR applications, where, instead of a classification problem, the input and output signals are sequential in nature. 

In this paper we propose to integrate this adversarial domain adaptation for the recognition of cursive handwriting recognition using an encoder-decoder framework. Thus, both the inputs and outputs of our system are variable-length signals formed by a sequence of characters. Although the same character set has to be used for both source and target domains, the proposed method is not restricted to a particular output lexicon nor language. We incorporate a temporal pooling step aimed at adjusting the adversarial domain adaptation techniques to problems having variable-length signals. To the best of our knowledge, just the recent parallel work of Zhang \etal~\cite{zhang2019sequence} proposes a similar idea. However, they propose that both the recognition and discrimination steps focus on character level. By disentangling the recognition and discrimination processes, one working at character and the other at word level respectively, we significantly outperform their approach. In addition, by synthetically rendering the source words with truetype fonts, our system does not require any manually generated label, and is trained ``for free'', not requiring any real annotated training data to be used as source domain.

\section{Adaptable Handwritten Word Recognition}
\subsection{Problem Formulation}

Our main objective is to propose an adaptable handwritten word recognizer application that is initially trained by synthetically generated word images, and then adapted to a specific handwriting style in an unsupervised and end-to-end manner. Our architecture, depicted in Fig.~\ref{fig:arch}, consists of two interconnected branches, the handwriting recognizer and the discriminator, in charge of the adaptation process. By means of a gradient reversal layer, the two blocks will play an adversarial game in order to obtain an intermediate feature representation that is indistinguishable whether it is generated from a real or synthetic input, while being representative enough to yield good transcription performances.

\begin{figure*}[h!tb]
%\begin{center}
\begin{tabular}{c}
\includegraphics[height=1cm]{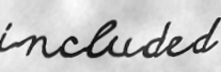}
\includegraphics[height=1cm]{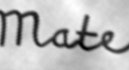}
\includegraphics[height=1cm]{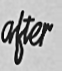}
\includegraphics[height=1cm]{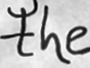}
\includegraphics[height=1cm]{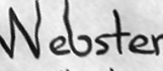}
\includegraphics[height=1cm]{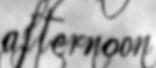}
\includegraphics[height=1cm]{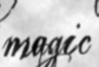}
\includegraphics[height=1cm]{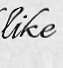}
\includegraphics[height=1cm]{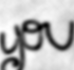}
\includegraphics[height=1cm]{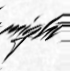}\\
\includegraphics[height=1cm]{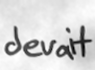}
\includegraphics[height=1cm]{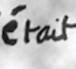}
\includegraphics[height=1cm]{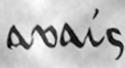}
\includegraphics[height=1cm]{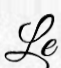}
\includegraphics[height=1cm]{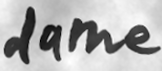}
\includegraphics[height=1cm]{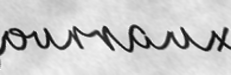}
\includegraphics[height=1cm]{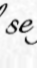}
\includegraphics[height=1cm]{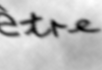}
\includegraphics[height=1cm]{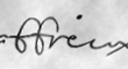}
\includegraphics[height=1cm]{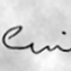}\\
\includegraphics[height=1cm]{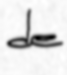}
\includegraphics[height=1cm]{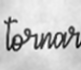}
\includegraphics[height=1cm]{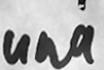}
\includegraphics[height=1cm]{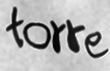}
\includegraphics[height=1cm]{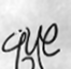}
\includegraphics[height=1cm]{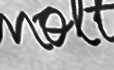}
\includegraphics[height=1cm]{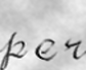}
\includegraphics[height=1cm]{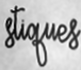}
\includegraphics[height=1cm]{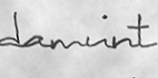}
\includegraphics[height=1cm]{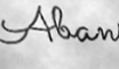}
\includegraphics[height=1cm]{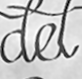}
\includegraphics[height=1cm]{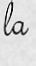}
\end{tabular}
\caption{Synthetically generated words in English (top), French (mid) and Catalan (bottom) used during training.}
\label{fig:syn_words}
\end{figure*}

In the proposed framework, two different flows are followed. The synthetically generated source words $x_i^s \in \mathcal{D}_s$ do come with their associated transcriptions $y_i^s \in \mathcal{Y}_s$, and enter both the recognizer and the discriminator branches. Contrary, the real target word images $x_i^t \in \mathcal{D}_t$, being unlabelled, are just processed through encoding and the discriminator block. At inference time, the prediction of the target texts $\hat{y}_i^t$ is not bounded by any previously defined lexicon, being totally independent of $\mathcal{Y}_s$.

\subsection{Rendering Synthetic Sources}

The use of synthetically generated word collections that look like real handwriting to magnify training data volumes has become a common practice. Although several public datasets, such as the IIIT-HWS dataset~\cite{krishnan2016generating}, exist, we decided to create our own, in order to include special characters (e.g. accents, umlauts, punctuation symbols, etc.) that we want our recognizer to tackle. 387 freely available electronic fonts that imitate cursive handwriting were selected. A text corpus consisting of over 430,000 unique words was collected from free ebooks written in English, French, Catalan and German languages. By randomly rendering those words with the different electronic fonts, we ended up with more than 5.6 million word images. In order to add more variability to the synthetic collection and to act as a regularizer, we incorporate a data augmentation step, specifically tailored to produce realistic deformations that one can find in handwritten data. This augmentation step is applied online within the data loader, so that each batch is randomly augmented. Pixel-level deformations include blurring, gamma, brightness and contrast adjustments or Gaussian noise. Geometric transformations such as shear, rotation, scaling and an elastic deformation are also randomly applied. Finally a model generating random background textures that simulate paper surface is applied. Some samples of synthetic words are shown in Fig.~\ref{fig:syn_words}. 

\subsection{Handwritten Word Recognition Framework}

We use an encoder-decoder architecture~\cite{bluche2017scan,sueiras2018offline,kang2018convolve,michael2019evaluating} topped with an attention mechanism as our handwritten word recognizer branch. Such architectures are able to process and output variable length data, and thus are not restricted to work with a predefined vocabulary.

\subsubsection{Encoder}

The aim of the encoder is to extract high-level features given a word image, which can be further adapted in the same feature hyperspace. In this work we define the encoder as a Convolutional Neural Network feature extractor followed by a Recurrent Neural Network. The initial CNN is in charge of extracting visual features that characterize the handwritten words. Concretely, the VGG-19-BN architecture~\cite{simonyan2014very} with pre-trained weights from ImageNet has been chosen. However, the classifier and the last max pooling layer have been removed to preserve spatial information and to tackle narrow feature representation of small elements, such as a single punctuation mark. The VGG network is followed by a multi-layered Bi-directional Gated Recurrent Unit (BGRU), which combines mutual information and extra positional information to the final feature representation $\mathcal{H}$. We denote $h_i \in \mathcal{H}, i \in \{1,2,...,N\}$ as the output sequence of the encoder. $N$ is the length of $\mathcal{H}$, which varies according to the lengths of the input word images. Thus, we denote $G_e:\mathcal{I}\rightarrow \mathbb{R}^{D\times N}$ as the encoder function given an image $I\in\mathcal{I}$ with parameters $\theta_e$.

\subsubsection{Attention-based Decoder}

The decoder is a one-directional multi-layered GRU, which predicts one character $\hat{y}_{i,k}^s$  at each time step $k$ until reaching the maximum number of steps $T$ or meeting the end of sequence symbol $\langle \operatorname{end} \rangle$. Thus, let $G_r$ denote the decoder function given the output of encoder $\mathcal{H}\in\mathbb{R}^{D\times N}$ with parameters $\theta_r$, and its output is a sequence of characters $\hat{y}_i^s$, which is the concatenation of $\hat{y}_{i,k}^s$, where $k \in \{1, 2, ..., T\}$.

In handwriting recognition, the attention should be ordered, for example, from left to right for germanic and roman languages. Although we have already applied BGRU in the encoder to add the positional information, we must give the attention a strong constraint: images should be read from left to right. For this reason, we have chosen the Location-based attention mechanism~\cite{chorowski2015attention}, because it takes into account the location information explicitly. At the current time step $k$, we extract $p$ vectors $l_{k,i}\in\mathbb{R}^p$ for every position $i$ of the previous attention mask $\alpha_{k-1}$ by convolving it with a matrix $F\in\mathbb{R}^{p\times r}$. Formally,
\begin{equation}
    l_k = F * \alpha_{k-1}.
\end{equation}
So we can obtain the attention mask $\alpha_k$ by
\begin{equation}
    \alpha_{k} = \operatorname{Softmax}(e_{k}), \label{equ:Softmax}
\end{equation}
where
\begin{align}
    e_{k, i} &= f'(h_{i}, s_{k-1}, l_{k})\nonumber\\
    &= w^T \tanh(Wh_{i} + Vs_{k-1} + Ul_{k,i}+ b),
\end{align}
where $w$, $W$, $V$, $U$ and $b$ are trainable parameters.

%%%%%%%%%%%%%%%%%%%%%%%%%%%%%%%%%%%%%%%%%%%%%%%%%%%%%%%%%%%%%%%%%%%%%%
%%%%%%%%%%%%%%%%%%%%%%%%%%%%%%%%%%%%%%%%%%%%%%%%%%%%%%%%%%%%%%%%%%%%%%

\subsection{Temporal Pooling for Unsupervised Writer Adaptation}
\label{subsec:domain_classifier}

Text being a sequential and variable-length signal, state-of-the-art adversarial domain adaptation methods can not be straightforwardly used, since they all rely on having fixed length feature vectors. We propose to explore several \emph{Temporal Pooling} strategies in order to transfer the variable length feature representation $\mathcal{H}$ into a fixed size feature representation $\mathcal{F}$ within the discriminator module:

\textbf{Column-wise Mean Value (CMV)} treats $\mathcal{H}$ as a column-wise sequence feature. The mean value is calculated as follows
\begin{equation}
    \mathcal{F} = \frac{1}{N}\sum_{i=1}^{N} h_i.
\end{equation}

\textbf{Spatial Pyramid Pooling (SPP)}~\cite{he2014spatial} is a flexible solution for handling different scales, sizes and aspect ratios of images. It severs the images into divisions from finer to coarser levels and aggregates local features in a fixed-size feature vector.

\textbf{Temporal Pyramid Pooling (TPP)}~\cite{wang2017temporal} is an one-directional SPP. It is considered to be more suitable for handwriting recognition tasks, because words are composed of a sequence of characters, and they are read in a specific direction.

\textbf{Gated Recurrent Unit (GRU)} are used to process the sequential signal $\mathcal{H}$ to output a fixed size feature representation $\mathcal{F}$. In our model, we simply apply a 2-layered one-directional GRU.

Once we have obtained a fixed representation, $\mathcal{F}$ is fed into the domain classifier, which consists of three fully connected layers with batch normalization and ReLU activation. $\theta_d$ is used to represent the parameters of the discriminator $G_d$. The output of $G_d$ is binary, either predicting that the features $\mathcal{F}$ come from source or target samples.

\subsection{Learning Objectives}
Until now, we have a recognition loss $L_r$ from the decoder and a discriminator loss $L_d$ from the domain classifier. Since our model is trained in end-to-end fashion, the overall loss for the training scheme is defined as

\begin{equation}
    \begin{aligned}
        L(\theta_e, \theta_r, \theta_d) = & \sum_{x_i\in \mathcal{D}_s} L_r\left(G_r\left(G_e(x_i)\right), y_i\right) - \\
        & \lambda \sum_{x_j\in \mathcal{D}_s\cup \mathcal{D}_t} L_d\left(G_d\left(G_e(x_j)\right), d_j\right),
    \end{aligned}
    \label{equ:loss}
\end{equation}
where $\lambda$ is a hyper-parameter to trade off the two losses $L_r$ and $L_d$. In Section~\ref{subsec:ablation}, different $\lambda$ methods have been studied.

As stated before, the source data consists in synthetic word images plus their corresponding labels. The target data corresponds to real word images, but without labels. 

The parameters of the discriminator are randomly initialized during the writer adaptation process. For the forward pass, the synthetic word images can be transferred through both the recognizer and the discriminator, while the real word images can only contribute to the discriminator loss. The backward propagation follows the same but reverse flow of the model by applying a \emph{Gradient Reversal Layer} (GRL)~\cite{ganin2014unsupervised} between the encoder and the discriminator. This layer applies the identity function during the forward pass but during the backward pass it multiplies the gradients by the parameter $-\lambda$. Thus, this layer reverses the gradient sign that flows through the model. By doing so, the model can be trained in a min-max optimization fashion. Minimizing the discriminator loss means to train a better discriminator for distinguishing between the synthetic and real data. In contrast, maximizing the discriminator loss for the encoder means to eliminate the differences of data feature distribution between the synthetic and real data. The goal of the optimization process is to find a saddle point that
\begin{equation}
        \hat{\theta}_e, \hat{\theta}_r = \argmin_{\theta_e, \theta_r} L(\theta_e, \theta_r, \theta_d)
    \label{equ:theta_er}
\end{equation}

\begin{equation}
        \hat{\theta}_d = \argmax_{\theta_d} L(\theta_e, \theta_r, \theta_d).
    \label{equ:theta_d}
\end{equation}

In short, synthetic data contributes to both the recognizer and the discriminator, whereas real data only contributes to the discriminator.

%%%%%%%%%%%%%%%%%%%%%%%%%%%%%%%%%%%%%%%%%%%%%%%%%%%%%%%%%%%%%%%%%%%%%%
%%%%%%%%%%%%%%%%%%%%%%%%%%%%%%%%%%%%%%%%%%%%%%%%%%%%%%%%%%%%%%%%%%%%%%

\section{Experiments}

\begin{table}
    \centering
        \caption{Overview of the different datasets used in this work depicting its characteristics.}
    \label{tab:dbs}
    \footnotesize
    \begin{tabular}{lllll}
    \toprule
    \textbf{Dataset} & \textbf{Words} & \textbf{Writers} & \textbf{Period} & \textbf{Language}\\
    \midrule
         GW~\cite{lavrenko2004holistic}     & 4,860 & 1 & Historic & English\\
         IAM~\cite{marti2002iam}    & 115,320 & 657 & Modern & English\\
         Rimes~\cite{augustin2006rimes}  & 66,978 & 1,300 & Modern  & French\\
         Esposalles~\cite{fornes2017icdar} & 39,527 & 1 & Historic & Catalan\\
         CVL~\cite{kleber2013cvl} &99,902 & 310& Modern & English/German\\ 
         \bottomrule
    \end{tabular}
\end{table}

In order to carry our writer adaptation experiments, we will use five different publicly available datasets with different particularities: single or multiple writers, coming from historic or modern documents or written in English, French, Catalan or German. We provide the details of such datasets in Table~\ref{tab:dbs}. To evaluate the system's performance, we will use the standard \emph{Character Error Rate} (CER) and \emph{Word Error Rate} (WER) metrics. In the tables, these values are in percentage ranging from [0-100].

\subsection{Implementation Details}

All our experiments were run using  PyTorch~\cite{paszke2017automatic} on a cluster of NVIDIA GPUs. The training was done using the Adam optimizer with an initial learning rate of $2 \cdot 10^{-4}$ and a batch size of 32. We have set the dropout probability to be 50\% for all the GRU layers except the last layer of both the encoder and decoder. During training, we have kept a balance in the total number of samples shown for both synthetic source words and real unlabelled target data. However, the training set is shuffled at each epoch and source and target data balancing is not guaranteed within a batch.

\subsection{Ablation Study}
\label{subsec:ablation}
Before assessing the performance of the proposed unsupervised writer adaptation model, we want to validate the adequacy of several hyper-parameters involved in our system. The following experiments are carried out using the IAM validation set as target dataset, except the last one, where the GW dataset was used instead. First, we evaluate which is the best temporal pooling strategy to recast the variable-length features of the encoder to the fixed-length features needed by the discriminator. In Table~\ref{tab:domain_classifier}, we observe that the GRU achieves the best performance. The GRU module has trainable parameters, so, contrary to the other aggregation strategies, it can learn how to effectively pool the variable-length features into a meaningful fixed-length representation, and consequently, obtain a better performance. For the rest of experiments we will use the GRU as our temporal pooling strategy.

\begin{table}[!htb]
    \centering
    \caption{Study on the different Temporal Pooling approaches of the discriminator, evaluated on the IAM validation set.}
    \label{tab:domain_classifier}
    \begin{tabular}{lcccc}
         \toprule
         & \textbf{CMV}  & \textbf{SPP} & \textbf{TPP} & \textbf{GRU} \\
         \midrule
         CER & 14.83 & 15.76 & 14.55 & \textbf{13.58}\\
         WER & 36.83 & 38.86 & 36.44 & \textbf{33.99}\\
         \bottomrule
    \end{tabular}
\end{table}

Second, we analyze three different approaches to set the hyper-parameter $\lambda$, which controls the trade-off between the recognition loss $L_r$ and the discriminator loss $L_d$. We choose to either set it as a constant $\lambda=1$, increase its value linearly from 0 to 1 at each epoch, or increase its value from 0 to 1 in an exponential way. Although a gradual increase of the weight of the discriminator loss could potentially benefit the overall performance, in Table~\ref{tab:lambda} we appreciate that simply setting $\lambda$ as a constant value provides the best results. 

\begin{table}[!htb]
    \centering
    \caption{Study on the different $\lambda$ strategies, evaluated on the IAM validation set.}
    \label{tab:lambda}
    \begin{tabular}{lccc}
         \toprule
        $\lambda$ & \textbf{Constant} & \textbf{Linear} & \textbf{Exponential}\\
         \midrule
         CER & \textbf{13.58} & 13.79 & 14.43 \\
         WER & \textbf{33.99} & 35.42 & 36.65 \\
         \bottomrule
    \end{tabular}

\end{table}

\begin{figure}[!htb]
%\begin{center}
\includegraphics[width=\columnwidth]{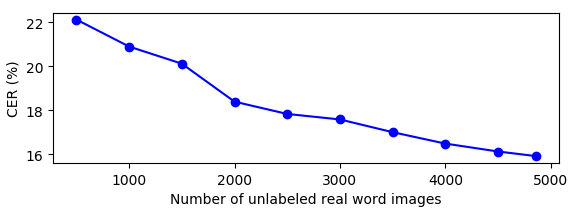}
%\end{center} 
   \caption{Influence of the amount of unlabeled real word images over the performance, evaluated on the GW dataset.}
\label{fig:curve}
\end{figure}

Finally, we explore the effect of providing different amounts of unlabelled target data to the system during writer adaptation. In this experiment we use the GW dataset, since it contains almost $5,000$ words from the same writer. We observe in Fig.~\ref{fig:curve} that the higher the amount of unlabelled target data, the lower the error rate. Thus, for the subsequent experiments, we will use all the available target data at hand during the adaptation, no matter if the scenario concerns a single writer or several of them. For multi-writer collections, we could thus choose among two options: (i) the system is adapted to a particular writer, using just a subset of the collection; (ii) the system is adapted to the whole collection style (rather than to the individual writing characteristics) by providing the whole dataset during the adaptation.

\subsection{From synthetic to real writer adaptation}
\label{subsec:res}

\begin{table*}[!htb]
    \centering
    \caption{Unsupervised writer adaptation results for handwritten word recognition. The gap reduction shows the improvement when the HTR, trained on synthetic data, is adapted to real data.}
    \label{tab:res_all}
    \begin{tabular}{lcccccccccccccc}
    \toprule
      & \multicolumn{2}{c}{\textbf{GW}} && \multicolumn{2}{c}{\textbf{IAM}} && \multicolumn{2}{c}{\textbf{Rimes}} && \multicolumn{2}{c}{\textbf{CVL}}&& \multicolumn{2}{c}{\textbf{Esposalles}}\\
    \cmidrule{2-3}\cmidrule{5-6}\cmidrule{8-9}\cmidrule{11-12}\cmidrule{14-15}
    & CER & WER && CER & WER && CER & WER && CER & WER && CER & WER\\
    \midrule
    Real target only & 4.56 & 13.49 && 6.88 & 17.45 && 2.80 & 8.51 && 3.64 & 7.77 && 0.47 & 1.68 \\
    Synth. source only & 26.05 & 56.79 && 26.44 & 54.56 && 21.46 & 52.48  && 26.30 & 55.64 && 30.78 & 66.33 \\
    Uns. adaptation & 16.28 & 39.95 && 14.05 & 34.86 && 14.39 & 39.21  && 19.19 & 44.29 && 20.96 & 50.00 \\
    \midrule
    Gap reduction (\%) & 45.46 & 38.89 && 63.34 & 53.09 && 37.89 & 30.18  && 31.38 & 23.71 && 32.40 & 25.26 \\
    %\midrule
    %Impr. percentage (\%) & 13.21 & 38.97 && 16.84 & 43.35 && 9.00 & 27.93 && 9.65 & 25.59 && 14.19 & 48.50\\
    \bottomrule
    \end{tabular}
\end{table*}

For the unsupervised writer adaptation experiments, we will use all the available images from each dataset during the unsupervised writer adaptation process, in order to have as much real word instances as possible. According to the experiments in the previous section, this should yield the best performance. It should be noted that these datasets are always used in an unsupervised manner, i.e. the system has access to the word images, but never to their transcriptions (labels). However, the CER and WER results are computed on the official test set partitions in all datasets, so that those results are comparable with the literature.

In Table~\ref{tab:res_all} we present our writer adaptation results on the five different datasets. For each dataset we also provide two baseline results. Training using target labels and training just using the synthetic samples provide baselines for the best and worst case scenarios respectively, either using ground-truth labels or ignoring any labelled information. The gap reduction is an measurement used to measure the effectiveness of the adaptation method which is defined as:
% \begin{equation}
%     gap\ reduction = \dfrac{error(synth.)-error(adapted)}{error(synth.)-error(real)}
% \end{equation}
\begin{equation}
    \operatorname{gap\ reduction} = \dfrac{\operatorname{error}(\operatorname{synth.})-\operatorname{error}(\operatorname{adapted})}{\operatorname{error}(\operatorname{synth.})-\operatorname{error}(\operatorname{real})}
\end{equation}
We appreciate that, in general, the difference in CER between these two baselines, lower bound $error(synth.)$ and upper bound $error(real)$, is about $20$ points, with the exception of the Esposalles dataset, which presents a much higher gap. This difference is most likely justified because it is the dataset in which the handwriting style differs more from a visual point of view from the synthetically generated samples. 

\begin{figure}[h]
     \centering
     \begin{tabular}{cc}
     \includegraphics[width=0.45\columnwidth]{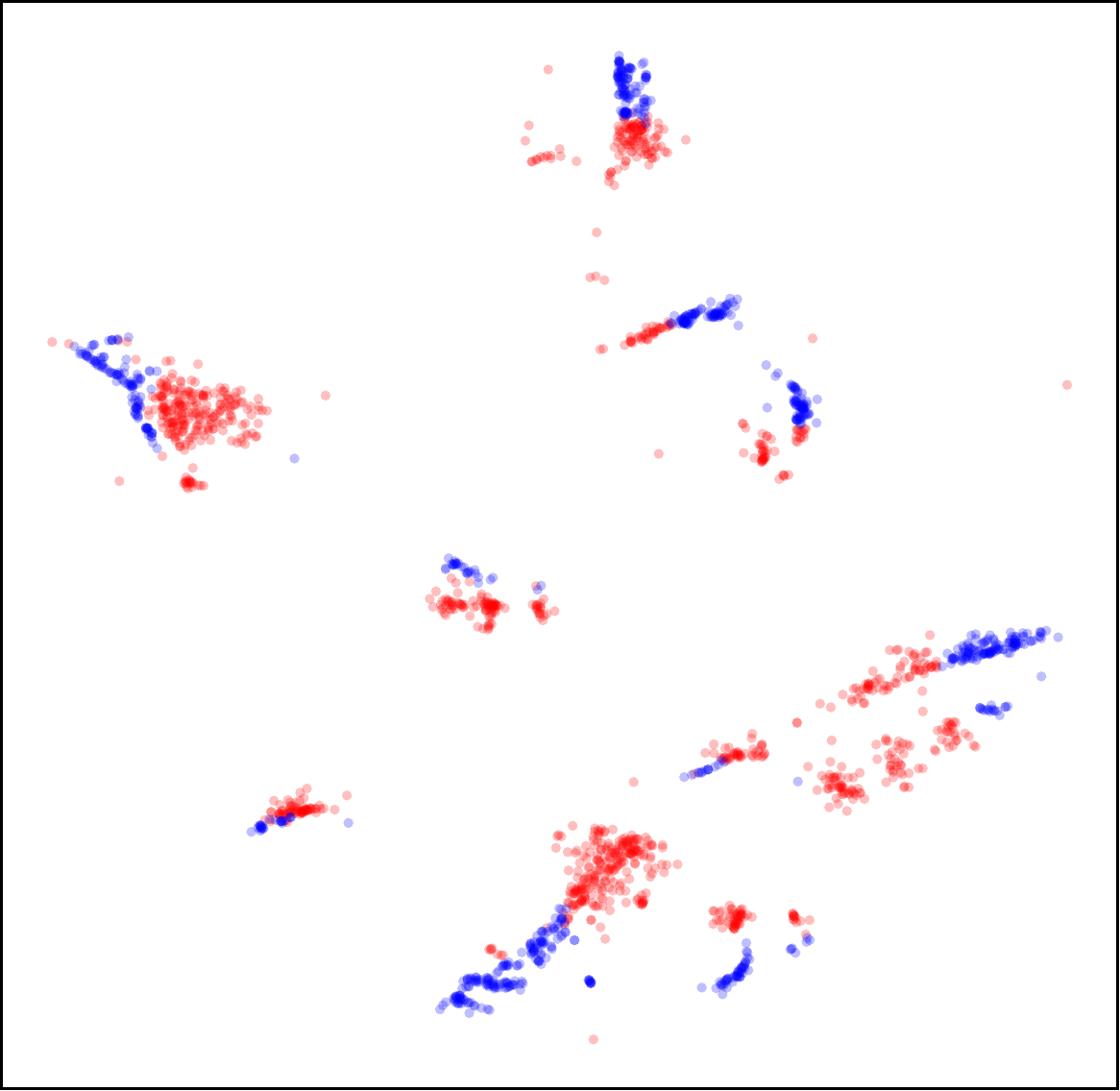}&
     \includegraphics[width=0.45\columnwidth]{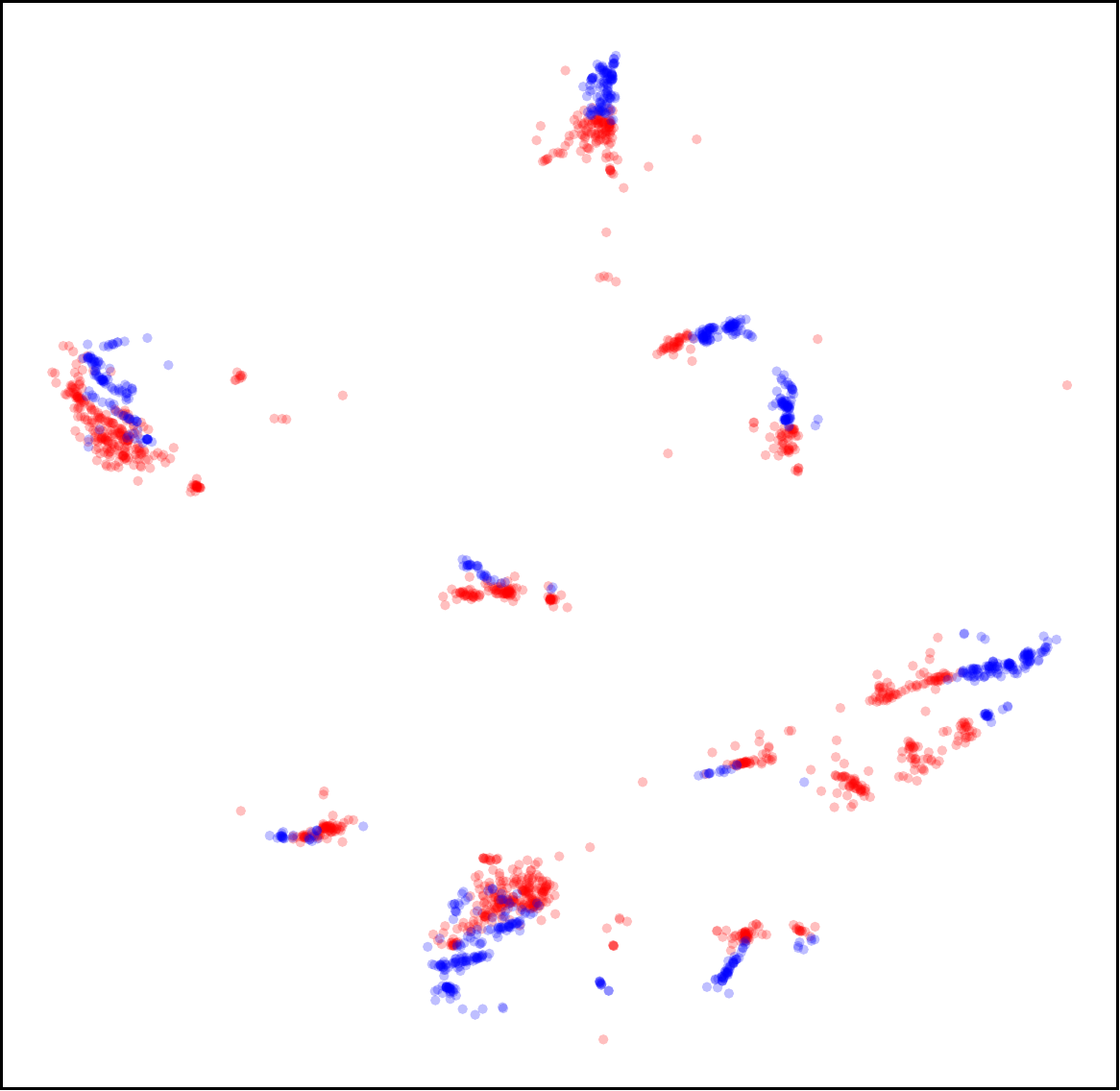}\\
     a) Before adaptation&b) After adaptation\\
     \end{tabular}
     \caption{The distribution of source (blue) and target (red) domain samples before (a) and after (b) the adaption to the GW dataset for the ten most common words.}
     \label{fig:davis}
 \end{figure}
 
Concerning the unsupervised writer adaptation results (in Table~\ref{tab:res_all}, \emph{Uns. adaptation}), we appreciate a significant improvement when compared with the sole use of synthetic training samples. The gap reduction ranges from $20\%$ in the worse case (CVL), up to $60\%$ in the best case (IAM). It is true that these results are worse than the ones obtained by a recognizer trained on labelled target data. However, the loss in accuracy is compensated by the fact that our approach is more generic and flexible: it is trained with synthetically generated data and it does not require any manually annotated target data for writer adaptation. In Fig.~\ref{fig:davis} we provide a tSNE visualization of the sample distribution before and after the unsupervised writer adaptation in the single-writer GW dataset.

\subsection{Writer adaptation with few samples}

This experiment is devised to evaluate whether the adaptation ability of our approach decreases when there are few samples in the target domain. Indeed, in the experiments presented in Table~\ref{tab:res_all}, the system is adapting to a particular individual handwriting style for the GW and Esposalles datasets, because they are single writer. Given that the IAM, Rimes and CVL datasets contain samples from multiple writers, the system is adapting from synthetic samples to the overall collection style. Since in the IAM dataset we do have groundtruth information about which specific writer produced each word, we choosed it for this writer specific adaptation experiment, taking into account that the volume of words per writer that we can use as target domain is very reduced. Within the IAM validation set, each writer has written between 13 and 602 words. As source domain we randomly selected 600 synthetic words (images and labels) for every single writer specific adaptation experiment. 

From the results shown in Table~\ref{tab:writer_adapt}, we appreciate that our model boosts the recognition performance on every writer even though when there is a very reduced amount of both source and unlabelled target samples. Due to the limited space, we only show the top five best and worse cases ranked by the improvement percentage between the CER measure obtained with a system trained with just synthetic data or after writer adaptation using this low amount of samples. We observe that for all the writers in the IAM validation set, the CER measure is enhanced after the proposed unsupervised adaptation. By inspecting the qualitative results, we observe that the writers that present the lowest improvement corresponded to specimens with writing styles that are visually very dissimilar to our synthetically generated source material. 

In general, this experiment depicts a realistic scenario in which our generic handwritten word recognizer, fully trained with synthetic data, is adapted to a new incoming writer by just providing a very reduced set of his handwriting. From the results, we can conclude that the recognition performance for this new writer is significantly boosted in most cases, in an unsupervised and efficient manner.

\begin{table}%[H]
    \centering
    \caption{Writer adaptation results, in terms of the CER, ranked by the improvement percentage with respect to the synthetic training.}
    \label{tab:writer_adapt}
    \begin{tabular}{lcccc}
         \toprule
         \textbf{Writer ID} & \textbf{Words} & \textbf{Synth.} & \textbf{Adapt.} & \textbf{Improv.}(\%) \\
         \midrule
         ID202 & 396 & 13.65 & 3.96 & 71.0 \\
         ID521 & 48 & 21.68 & 7.39 & 65.9 \\
         ID278 & 129 & 7.71 & 3.66 & 52.5 \\
         ID625 & 80 & 23.18 & 11.76 & 49.3 \\
         ID210 & 136 & 9.41 & 5.29 & 43.8 \\
         \ldots & \ldots & \ldots & \ldots & \ldots \\
         ID533 & 52 & 37.50 & 32.50 & 13.3 \\
         ID182 & 69 & 29.89 & 26.05 & 12.8 \\
         ID515 & 74 & 38.29 & 34.20 & 10.7 \\
         ID527 & 127 & 24.86 & 22.20 & 10.7 \\
         ID612 & 55 & 29.83 & 28.57 & 4.2 \\
         \midrule
         Mean & 135 & 24.32 & 18.34 & 27.4 \\
         \bottomrule
    \end{tabular}
\end{table}

\subsection{Comparison with the state of the art}

\myparagraph{Supervised fine-tuning}. In order to put into context our reached results, we compare in Table~\ref{tab:finetune} our model with the state-of-the-art approaches that propose to pre-train a handwriting recognizer with a large dataset, e.g. IAM, and then fine-tune the network to transfer the learned parameters to a different collection, e.g. GW, with a disparate style. We compare against the recent works proposed by Nair \etal~\cite{nair2018knowledge} and Arandillas \etal~\cite{aradillas2018boosting}. They achieve CER values of 59.3\% and 82\%, respectively with their models trained on IAM and tested over the GW test set. Our baseline model, pre-trained just using a synthetically produced data, already achieves a 26.05\% CER on the GW dataset. This backs up the intuition that the use of a synthetic dataset, which can contain as many training samples as desired, provides better generalization than training with a much shorter amount of real data. 

Our unsupervised writer adaptation reaches a 16.28\% CER while Nair \etal and Arandillas \etal reach a 8.26\% and 5.3\% CER respectively when fine-tuning, at the expense of requiring a fair amount of manually labeled data. Obviously, our unsupervised approach does not reach the same performance as these supervised approaches, because they use labelled GW words. Although it is not the main scope of our paper, if we do use labels for the target domain (last row in Table~\ref{tab:finetune}), i.e. we adapt to the new incoming writer in a supervised manner, our approach outperforms the above methods, reaching a 2.99\% CER.

\myparagraph{Unsupervised domain adaptation}. To the best of our knowledge, only the work of Zhang \etal~\cite{zhang2019sequence} report results for unsupervised writer adaptation at word level. However, for the case of handwriting words, they propose to use labelled IAM training data as source and unlabelled IAM test data as target domains. In our opinion, such experiment does not present any significant domain shift. When using their same experimental setting, shown in Table~\ref{tab:state-of-the-art}, our approach achieves a significant better performance.

\begin{table}[!tb]
    \centering
    \caption{Comparison with supervised fine-tuning.}
    \label{tab:finetune}
    \begin{tabular}{lllc}
    \toprule
    \textbf{Method} & \textbf{Train} & \begin{tabular}{@{}l@{}}\textbf{Fine-tuning} \\ \textbf{adaptation}\end{tabular} & \textbf{CER}\\
    \midrule
    \multirow{2}{*}{Nair~\cite{nair2018knowledge}}  & IAM  & None &59.30\\
                                & IAM & Sup. GW   & 8.26\\
    \midrule
    \multirow{2}{*}{Arandillas~\cite{aradillas2018boosting}} & IAM &None& 82.00\\
                                    & IAM & Sup. GW & 5.30\\
    \midrule
     \multirow{3}{*}{Proposed} & Synth. &None& 26.05\\
    & Synth. &Uns. GW& 16.28\\
    & Synth. &Sup. GW& 2.99\\
    \bottomrule
    \end{tabular}
\end{table}

\begin{table}[!tb]
    \centering
    \caption{Comparison with sequence-to-sequence domain adaptation on IAM dataset.}
    \label{tab:state-of-the-art}
    \begin{tabular}{lccc}
    \toprule
    \textbf{Method} & \textbf{CER} &\textbf{WER} & \textbf{Average}\\
    \midrule
    Zhang~\etal~\cite{zhang2019sequence} & 8.50 & 22.20 & 15.35\\
    \textbf{Proposed} & \textbf{6.75} & \textbf{17.26} & \textbf{12.01}\\
    \bottomrule
    \end{tabular}
\end{table}

%%%%%%%%%%%%%%%%%%%%%%%%%%%%%%%%%%%%%%%%%%%%%%%%%%%%%%%%%%%%%%%%%%%%%%
%%%%%%%%%%%%%%%%%%%%%%%%%%%%%%%%%%%%%%%%%%%%%%%%%%%%%%%%%%%%%%%%%%%%%%
\section{Conclusion}
We have proposed a novel unsupervised writer adaptation application for handwritten text recognition. Our method is able to adapt a generic HTR model, trained only with synthetic data, towards real handwritten data in a completely unsupervised way. The system mutually makes the high-level feature distribution of synthetic and real handwritten words align towards each other, while training the recognizer with this common feature distribution. 

Our approach has shown very good performance on different datasets, including modern, historical, single and multi-writer document collections. Even when compared to supervised approaches, our approach demonstrates competitive results. Moreover, since our unsupervised approach only requires to have access to a few amount of word images from the target domain, but not their labels, we believe that it is a promising direction towards a universal HTR for unconstrained scenarios, e.g. industrial applications.

\section*{Acknowledgements}

This work has been partially supported by the European Commission H2020 SME Instrument program, project OMNIUS: SaaS platform for automated categorization and mapping of digitized documents using machine intelligence semantic extraction, grant number 849628, the grant 2016-DI-087 from the Secretaria d'Universitats i Recerca del Departament d'Economia i Coneixement de la Generalitat de Catalunya, TIN2017-89779-P, RTI2018-095645-B-C21, FPU15/06264 and RYC-2014-16831. The Titan XP used for this research were donated by NVIDIA.

{\small
\bibliographystyle{ieee}
\bibliography{egbib}
}

\end{document}